\documentclass[letterpaper, 10 pt, conference]{ieeeconf}
\IEEEoverridecommandlockouts
\usepackage{cite}
\usepackage{amsmath,amssymb,amsfonts}
\usepackage{algorithmic}
\usepackage{graphicx}
\usepackage{textcomp}
\usepackage{xcolor}
\usepackage{booktabs}
\usepackage{float}
\usepackage{fontawesome5}
\usepackage{hyperref}
\usepackage{cleveref}
\usepackage{listings}

\usepackage{xcolor}
\usepackage{siunitx}
\lstdefinestyle{prompt}{
basicstyle=\ttfamily\footnotesize,
breaklines=true,
breakindent=0pt,
columns=fullflexible,
keepspaces=true,
frame=single,
framesep=4pt,
rulecolor=\color{black!50},
backgroundcolor=\color{black!3},
xleftmargin=4pt, xrightmargin=4pt,
aboveskip=6pt, belowskip=6pt,
} 
\usepackage{subcaption}
\usepackage{pgfplots}
\pgfplotsset{compat=1.17}
\usepgfplotslibrary{fillbetween}  \usepgfplotslibrary{patchplots}   
\usepackage{tikz}
\usetikzlibrary{positioning, arrows.meta, shapes.geometric, fit, backgrounds,calc}
\definecolor{accent}{RGB}{0,90,140}
\definecolor{soft}{RGB}{240,243,247}
\definecolor{simfill}{RGB}{226,232,240}
\def\BibTeX{{\rm B\kern-.05em{\sc i\kern-.025em b}\kern-.08em
    T\kern-.1667em\lower.7ex\hbox{E}\kern-.125emX}}

\begin{document}

\title{\LARGE \bf{Large Language Model Explainability for \\
Planning Lunar Rover Trajectories under Uncertainty
}}

\author{Michelle Ho$^{1}$, Maximilian Adang$^{1}$, Zachary Booth$^{2}$, David E. Astle$^{2}$, \\ Margaret Hansen$^{4}$, Edward Balaban$^{3}$, and Mykel J. Kochenderfer$^{1}$
\thanks{$^{1}$Stanford University, Stanford, CA 94305
{\tt\small mtho@stanford.edu}}%
\thanks{$^{2}$ KBR Inc., $^{3}$Intelligent Systems Division, NASA Ames Research Center, Moffett
Field, CA, 94035}
\thanks{$^{4}$Carnegie Mellon University, Pittsburgh, PA 15213}%
}

\maketitle

\begin{abstract}
 Planning planetary rover trajectories under probabilistic uncertainty is a multi-objective optimization problem that science teams currently iteratively solve and verify by hand. This process becomes impractical when timelines and constraints frequently shift throughout the mission planning phase. Sampling-based planners such as Monte Carlo tree search can speed up this process, but their outputs do not preserve traceability to the science objectives that justify each plan, which the science team needs to verify. We propose an interactive assistant based on a large language model (LLM) that can explain planner trajectories using domain vocabulary and re-run the planner under user-specified conditions to compare alternatives. Previous work has used LLMs inside planner decision-making loops, but we explore using them for post-planning explanation, where they are more likely to be trusted since they do not directly influence planner decisions. Our framework pairs a simulator inspired by the NASA Volatiles Investigating Polar Exploration Rover (VIPER) mission with a ReAct queryable interface that calls curated action tools and retrieval functions. We evaluate the assistant on a query bank consisting of plan comparisons, feasibility, and lookup queries. Experiments show that 90.9\% of the assistant's responses are faithful to the planner's outputs, a nearly 65\% improvement over zero-shot prompting of the same LLM without tool access, demonstrating the potential for trustworthy explainability of uncertainty-aware traverses with an LLM-based system.

\end{abstract}

\section{Introduction} 

For space missions, planning the science phase typically requires vehicle trajectory optimization for multiple objectives and constraints. These trajectories also depend on many sources of probabilistic uncertainty, which become impractical to reiterate on quickly and frequently by hand. A simulator and an accompanying automated planning algorithm can speed up this iteration process, but their outputs may not be directly useful to the scientists who need to map these outputs to the effects on science data collection. This gap in explainability motivates an automated approach that uses large language models (LLMs) to explain trajectories in a comprehensible form to users who may not be familiar with the inner workings of the planning system.

Advances in robotics have been driven by LLMs applied to motion planning for tasks with high-dimensional observations, multi-modal sequences and behaviors, and complex failure modes \cite{ding2023task, chen2024autotamp, wang2024llm3}. For instance, recent work uses LLMs for planning Martian rover traverses \cite{nasa2025perseveranceai}, and efforts have begun to integrate LLMs with structured planners as state trackers or plan verifiers within the planning stage \cite{chen2023llmstate, zhao2023large}. Improvements to LLM in-context reasoning, text generation, and information retrieval \cite{zhao2023survey} suggest that they have the skills to provide comprehensive explainability for these tasks. However, current work does not use LLM reasoning capabilities to explain long-horizon plans made by rigorous techniques for planning under uncertainty. 

One application that could benefit from this approach is planning planetary rover traverses. Such missions must account for a fixed time budget, navigation to sites of high scientific value, and uncertain speeds. We are inspired by NASA's Volatiles Investigating Polar Exploration Rover (VIPER), which must explore a predefined area of high scientific value, called a science station, within a roughly 24-hour period. The rover must collect measurements of volatiles with its three continuously operating spectrometers (neutron, near-infrared, and mass) through prospecting and obtain subsurface regolith samples with its onboard drill \cite{colaprete2019overview}. The rover is solar-powered and thus needs to remain in sunlight as much as possible, unless operating within a permanently shadowed region (PSR) science station. The rover's speed also depends on the sun's position, as driving in certain directions is difficult or impossible due to the low-rising sun creating imaging issues for the navigation and hazard-avoidance cameras. The rover must also avoid communication shadows and slopes. Verifying by hand whether a handcrafted traverse satisfies all of these objectives and constraints is likely impractical, especially considering probabilistic uncertainties. An interpretable, automated planner is essential when timelines, constraints, and operational windows shift and require frequent replanning and verification. 

  \begin{figure}[t]
      \centering
      \includegraphics[width=0.99\columnwidth]{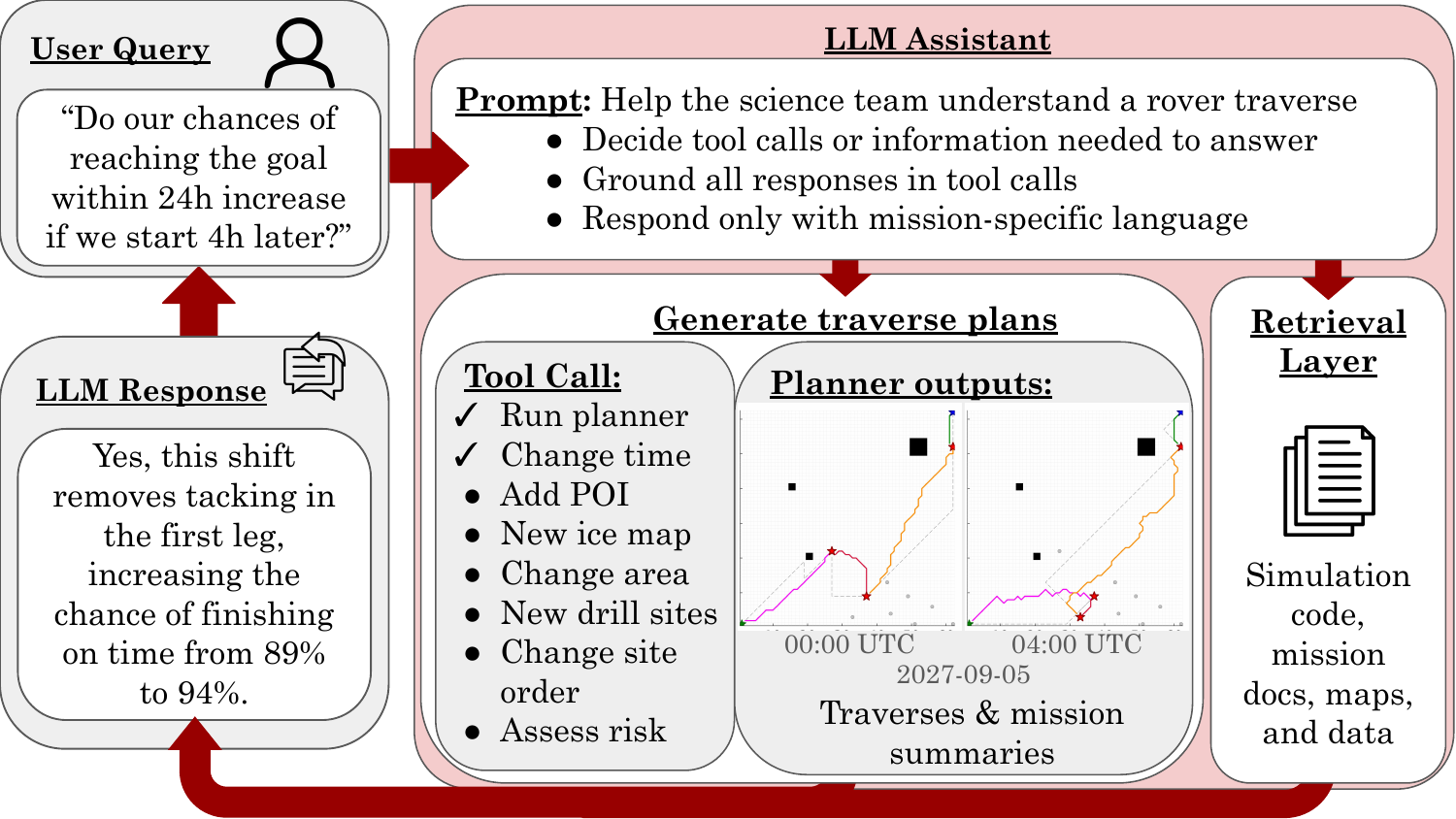}
      \caption{End-to-end flow of our LLM-based science station planning assistant
      on an example query}
      \label{fig:hero}
      \vspace{-0.7cm}
  \end{figure}

NASA's current tool for overseeing health-aware mission planning is the System Health-Enabled Real-time Planning Advisor (SHERPA) \cite{Balaban2025SHERPA} is a strategic mission planning AI system in use by NASA. SHERPA uses sampling-based \ methods like Monte Carlo Tree Search (MCTS) \cite{coulom2007efficient, kochenderfer2022algorithms} to compute and evaluate potential traverses for a variety of missions, such as VIPER's long-range traverses. However, it currently does not have support for the detailed planning inside VIPER's science stations. Additionally, MCTS plans trajectories by maximizing future rewards and rollouts, while the science team verifies trajectories against mission-specific objectives, such as risk of finishing over the time limit or maximizing prospecting area. The planner's outputs do not directly trace back to these objectives, so it may require additional analysis to determine whether a given trajectory satisfies them.

To address this gap, we make the following contributions towards producing explainable trajectories: 
\begin{enumerate}
    \item An MCTS-based optimal trajectory planner coupled with a VIPER-inspired, physics-based simulator. Inputs to the planner include ice stability depth maps and sun-azimuth-dependent probability distributions for rover speeds.
    \item An LLM-based, interactive assistant that dispatches the planner and a set of tools to automate replanning and explanation of trajectories with mission context, evaluated on a bank of science-team-inspired queries.
    
\end{enumerate}

The proposed framework seen in \Cref{fig:hero} is designed to allow the science team to efficiently replan under different conditions and constraints, while building trust with the trajectory planning algorithms, by keeping the team engaged in an interactive planning loop and grounding explanations in a familiar vocabulary. The LLM assistant does not override the planner, but rather runs it on behalf of the user with their specified conditions or explains resulting trajectories, without telling the user what they should do.
\section{Related Work}

\subsection{Sequential Decision Making under Uncertainty}

Trajectory planning under uncertainty is often formulated as a Markov decision process (MDP) \cite{kochenderfer2022algorithms}. Since exact solutions to large MDPs are computationally intractable, online sampling-based methods are often used, such as Monte Carlo Tree Search (MCTS) \cite{coulom2007efficient}, which builds a search tree of value estimates from simulated rollouts. We adopt a standard MDP formulation with an MCTS planner, as SHERPA does for strategic mission planning \cite{Balaban2025SHERPA}.

\subsection{Rover Trajectory Planning}

Rover planning has historically been a human-driven process. Gaines et al. \cite{gaines2016productivity} document the operational workflow for Mars rovers, where science teams hand-craft candidate traverses and validate them against constraints. They identify the cost of this loop as a primary limit on mission productivity. The process becomes even more complex in lunar polar missions, where a rover needs to avoid large, fast-moving illumination and communication shadows and where its speed, among other factors, depends on the difference between the heading and the azimuth to the sun. The latter consideration is due to the sun, which does not rise more than a few degrees above the horizon, creating imaging issues for the rover's navigation and hazard-avoidance cameras when traveling in certain directions. Another consideration in traverse planning is the achievement of the mission's scientific objectives (e.g., detecting and characterizing deposits of water ice or other volatile materials and maximizing exploration within the area). To incorporate such constraints and considerations, NASA uses SHERPA for the VIPER mission \cite{Balaban2025SHERPA}, which synthesizes mission plans that optimize scientific productivity and maximize robustness to delays, malfunctions, and other adverse conditions using sampling-based solvers such as MCTS \cite{coulom2007efficient, kochenderfer2022algorithms}. Prior automated lunar planners, in contrast, focused on safe reachability and topology and global path optimization \cite{lamarre2024safe,richter2024multiobjective,candela2022approach}. However, SHERPA's algorithmic framing may not be directly useful to the science team, who want to understand the solution within the context of science objectives. Our work targets this gap by using an LLM to explain the output from an MCTS-based planner that, coupled with a VIPER-inspired simulator, produces optimized rover trajectories.

\subsection{LLMs for Planning}

Previous work that integrates LLMs with planning has used them to define goals for task and motion planning \cite{ding2023task}, translate natural language task specifications into formal representations that planners consume \cite{chen2024autotamp}, reason about motion failures and propose corrections \cite{wang2024llm3}, and serve as state trackers for long-horizon planners \cite{chen2023llmstate}. Additionally, recent work has placed LLMs inside MCTS itself, using them as a heuristic to guide the search \cite{zhao2023large}. LLMs in these algorithms drive planners within the decision loop to produce better trajectories, rather than being used to explain the resulting output trajectories after the planner is done. 
 
One work has also used LLMs to explain decisions in dynamic environments, such as lane changing \cite{peng2024lc-llm}, but is limited to vehicle interaction modeling and does not consider using LLMs with motion planning algorithms. Other works explore dynamic reasoning for trajectories \cite{zhang2024unveiling} or combining MCTS with LLMs for self-training, evaluating intermediate reasoning steps for LLM outputs \cite{zhang2024rest-mcts}. These approaches are promising for interfacing LLMs with MCTS but are directed at improving LLMs rather than using them to explain trajectories. Another application merges LLM reasoning with MCTS planning for dynamic behaviors in video games, where LLMs and MCTS work together to create emergent narratives \cite{guignard2024exploring}. While relevant to interfacing LLMs and MCTS, this work aims to improve agent behavior with LLMs, instead of using them for explainability. 
  
\subsection{Grounding and Tool Use for LLMs}
Pretrained LLMs often perform worse on domain-specific tasks. Their broad training priors typically do not contain domain expert knowledge, so with zero-shot prompting, the model fills the gap with plausible-sounding but incorrect claims from its prior knowledge. Therefore, LLMs benefit from additional structure and knowledge for these tasks. For instance, retrieval-augmented generation (RAG) \cite{lewis2020rag} is a framework that grounds LLM responses in external data from a trusted source when queried. Additionally, tool-use frameworks treat the LLM as a controller that calls previously designed functions to inspect state or invoke external systems. ReAct (Reasoning and Acting framework) \cite{yao2023react} allows the LLM to perform both reasoning and actionable tool calls. DSPy \cite{khattab2024dspy} puts this framework into a programmable module, so rather than relying on exact keyword matches for retrieval, the LLM decides which tools to call from declarative signatures. Unlike previous applications of these frameworks individually, our setting requires both in different roles. Retrieval without action tools could only describe prior runs, while tool access without retrieval could not translate between mission-language queries and planner outputs.

\subsection{Explainable Planning}
With the steady improvements in LLMs, there is existing work that also considers using them to build trustworthiness for AI-based planning.
Previous work uses them to explain how automated planners justify decisions to humans through plan explanations, contrastive ``why not'' queries, and model reconciliation \cite{chakraborti2020emerging}. These methods are tied to planners whose actions and states are discrete and labeled. They do not transfer to trajectory planners that operate over continuous, nondeterministic dynamics like the MCTS solver we use. Some prior work has also involved translating robot control policies into natural-language descriptions \cite{hayes2017improving}, having robots verbalize their plans before acting \cite{rosenthal2016verbalization}, and providing causal model grounded explanations for the behavior of model-free reinforcement learning agents \cite{madumal2020explainable}. However, though these efforts are meant to build trust between humans and autonomous systems, they explain the planner's internal policy to a passive observer rather than the planner's outputs to a domain expert without an algorithmic background.

\section{Methodology}
We formulate our VIPER-inspired trajectory planning problem within a science station as a sequential decision-making problem, and develop an MCTS-based planner that uses a simulator grounded in mission requirements and constraints. We then develop an LLM-based framework for providing accurate explanations of the planning results with a retrieval layer and tool catalog for a ReAct module.

\subsection{Problem Formulation}\label{sec:problem_formulation}

We consider a rover that collects science data with contact measurements at a set number of drill sites and non-contact prospecting while traversing between the sites. The rover must complete all drills and reach a goal location before the station exit deadline, while also moving with an appropriate sun-azimuth-dependent speed made good (SMG), avoiding obstacles such as craters, and prospecting for at least 190 m. 

Drill site selection and planning prospecting segments are handled in two stages. The set of sites to be visited consists of the points of interest (POIs) given by the science team and the set number of drill sites that the planner selects. This selection considers a synthetic Gaussian process-based map that encodes information of interest. That information in this example is the scientific value of the ice volatiles. For a given location in the science station, each cell has an associated mean and variance \cite{hansen2026learning}. This map structure is consistent with previous work that represents maps of natural phenomena for informative path planning as Gaussian processes \cite{brouwer_informative_2021, hitz_adaptive_2017, candela_planetary_2017}. 

We choose $N_c{=}10$ candidate drill sites on the map by a weighted sum of the mean and variance of the scientific value of the volatiles, balancing exploring sites with more uncertainty with sites that provide the highest scientific value. We use a minimum spacing constraint to spread them out over the map. We down-select to $K{=}3$ drill sites by enumerating all $\binom{N_c}{K} \cdot K!$ orderings and picking the lowest-expected-time feasible ordering under deterministic speed, giving the largest exploration budget for the robot to perform the most prospecting while traversing to drill sites.  

We formulate the problem of prospecting and navigating between sites under probabilistic speeds as a Markov decision process (MDP), represented by the tuple $ (\mathcal{S}, \mathcal{A}, T, R, \gamma)$ \cite{kochenderfer2022algorithms}. Rather than use a large state space and select actions to move between individual cells, we treat this as a semi-MDP (SMDP)\cite{baykalsemi2007}, where we plan between nodes (start, goal, POIs, and drill sites) on a waypoint graph, $\mathcal{G}$. Thus, MCTS selects macro-actions, or which sites to visit in which order, and then we plan with A* to navigate in the cardinal and intercardinal directions through the cells between the nodes. The state $s$ is the tuple $(n, \mathbf{d}, k, \ell, t)$ collecting the rover's current waypoint node $n$, vector $\mathbf{d}$ with count $k$ of visited sites, total path steps $\ell$, and elapsed science station time $t$. 

The transition dynamics function $T(s' \mid s, a)$ is nondeterministic, since the elapsed time depends on sampling the SMG distribution along the A* path between nodes. This models the various sources of execution uncertainty, such as localization and navigation errors, communication delays, wheel slip, or temporary malfunctions. The SMG Gaussian distribution is clipped such that the sampled SMG never exceeds the provided deterministic best-case value. 

The SMG at each cell depends on the rover's heading relative to the sun azimuth $\phi(t)$:  
\begin{equation}
\Delta t \sim p_{\text{SMG}}\!\left(\cdot \mid \theta_{\text{rover}}(\text{s}),  \phi(t)\right),
\label{eq:smg}
\end{equation}
where $\Delta t$ is the time required to traverse to the next cell, $\theta_{\text{rover}}(s)$ is the rover's heading at state $s$, $\phi(t)$ is the sun's azimuth at science station time $t$ (obtained from a NASA ephemeris), and $p_{\text{SMG}}(\cdot \mid \theta_{\text{rover}}, \phi)$ is the empirical SMG distribution. Direct sunward or sundown travel requires tacking, or alternating between two oblique headings symmetric about the sunward direction due to the orientation of navigation and hazard-avoidance cameras. Turns add $\Delta t_{\text{turn}}=107$s to the elapsed time if they exceed a certain cone where the rover requires new images for the hazard map generation. 

The reward function is defined as
\begin{equation}
R = R_{\text{event}} - \alpha\,\Delta t - \beta\, \tau^2,
\label{eq:reward}
\end{equation}
where $R_{\text{event}}$ is a reward for static events at $s'$ (e.g., drilling a site, prospecting $\geq 190$ m, or reaching the goal). The time-rate term $\Delta t$ penalizes traverse time on the required route. Minimizing the time required directly maximizes the time available for prospecting, resulting in higher measurements coverage. The clipped fraction of the science station time budget elapsed at the next state $s'$ is $\tau = \min\{t' /T_{\text{budget}}, 1.5\}$, squared and clipped so the penalty grows sharply while approaching the limit, but does not penalize further for going over. The weights for the time-related terms are $\alpha, \beta > 0$. Each drill operation takes $5.5$ hours, leaving $\sim 7.5$ hours for traversing to the goal and prospecting in a 24-hour science station. Episodes terminate on completion success or time budget exhaustion, and we discount future rewards by $\gamma = 0.9$. Any craters or rocks are treated as obstacles. We solve this MDP with MCTS.

\subsection{Science Station Planner}\label{sec:simulator}

We instantiate the planner over an $n \times n$ grid with 1-m resolution. The maximum possible SMG is dependent on the sun's azimuth, and up-sun or down-sun travel approaches zero unless the rover tacks. Pairwise A* paths between waypoints are precomputed once per problem instance. After MCTS returns the lowest-expected-time traverse, a post-processing stage greedily inserts prospecting detours into each waypoint-to-waypoint segment, using the slack between the conservative P90 (90th percentile) traverse time and the science station budget to explore as much of the science station as is allowed. The planner outputs the chosen drill ordering, the enhanced trajectory, timing risk estimates, the extra prospected area, and the MCTS rollouts. This data is provided to the LLM assistant to provide comprehensive explanations to the user. 

\begin{figure}[!htbp]
      \centering
       \vspace{-0.3cm}
      \includegraphics[width=0.6\columnwidth]{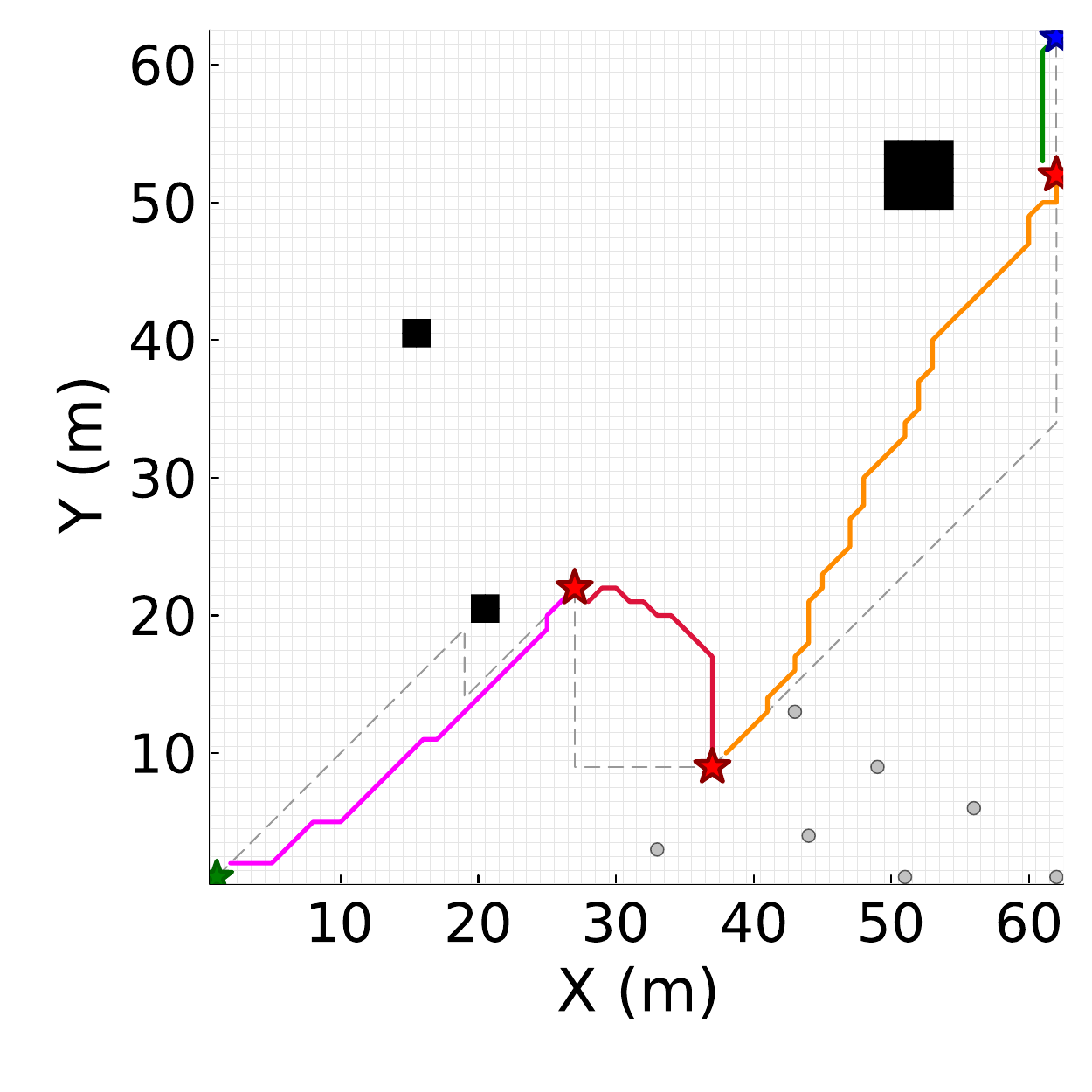}
      \vspace{-0.2cm}
      \caption{Example MCTS SST for a science station
      within a 62 m x 62 m square. The planner picks 3 drill sites (red stars) from 10 candidates (gray circles), routes a path from the start (green star) through each drill, and to the goal (blue star). The most direct path (gray, dashed) is shown for reference. Each drill-segment (pink, red, orange) has prospecting detours injected with the remaining time budget. The segment to the goal is in green. Black squares mark untraversable terrain.} 
      \label{fig:methodology-traj}
      \vspace{-0.3cm}
  \end{figure}

\subsection{LLM Assistant} \label{sec:assistant}

As seen in \Cref{fig:hero}, the assistant consists of three components that work with the planner: a chatbot user interface, a retrieval layer that grounds explanations in source material, and a tool catalog that the LLM calls. The science team interacts with the assistant through a chat window where queries are typed and sent to the LLM agent. The framework parses the query and responds back in the chat window. 

The retrieval layer gives the LLM the mission documents, planner source code (or the data), plots, and solutions that result from the planner as context when answering the user. Mission-document retrieval is implemented as a semantic search over chunked embeddings of the document. The source code and data retrieval are path lookups, so every cited fact is traceable to a specific file at a specific revision.

The chatbot is implemented with the ReAct framework \cite{yao2023react} as a programmable DSPy module, allowing for more robust reasoning and acting than zero-shot prompting or retrieval alone can achieve. This structure requires a tool catalog that provides actions for the LLM to call (which align with possible operations that the science team may want to perform), listed here:  

\begin{itemize}
      \item \textbf{Run science station}: Plan a full science station mission from scratch and return the summary.
      \item \textbf{Set start time}: Change the science station start date and time and replan, but reject start times whose 24-hour plan would result in sun shadows covering part or the entirety of the science station.
      \item \textbf{Add or replace POI}: Evaluate adding or replacing points of interest at specific coordinates. Returns feasibility and time impact.
      \item \textbf{Replan with new map}: Replan the science station using a different ice-map file.
      \item \textbf{Plan arbitrary region}: Plan a science station inside an arbitrarily shaped science region whose boundaries are defined by vertices given by the user.
      \item \textbf{Select custom drill sites}: Plan a science station using user-chosen drill sites from the candidate set. If infeasible, it suggests single-site swaps.
      \item \textbf{Force site order}: Replan with a user-specified site ordering rather than the planner's chosen order.
      \item \textbf{Risk assessment}: Roll out the current plan $N$ times under randomized, user-fixed, best-case, or worst-case speed uncertainty. Report the timing distribution (P50, P75, P90, P95, P99) and the success rate of reaching the goal end location. A success is defined as reaching the three drill sites before the goal within 24 hours and prospecting at least 190 m.
  \end{itemize}

These actions allow the user to quickly replan with new conditions, make comparisons between potential plans, or improve understanding of the decisions made by the planner. 

The LLM is instantiated with a prompt that takes the query as input and requires a natural-language response output. The LLM uses this prompt and the query to decide which tools and documents to call to generate the best response. The prompt to the LLM includes high-level guardrails to encourage responses that may resonate better with the science team. It constrains outputs to the operator vocabulary used by the science team (e.g., science station time budget, estimated ice quality, sun-relative SMG) and forbids algorithmic terminology (future rollouts, value estimates, reward functions). It also enforces that every factual claim in the response must trace either to a tool result or to the user query, not the model's prior knowledge. The LLM assistant does not modify the underlying MDP or override the planner. It can only run the planner on behalf of the user with their specified conditions or explain the results, without telling the user what they should do. 

\subsection{Experiments} \label{sec:experiments}

We design experiments to highlight the assistant's contributions: (i) does the assistant correctly parse the user's mission-language requests to identify the correct tools and information needed, and (ii) are the resulting explanations faithful to the planner's outputs and to the source material? We report multiple metrics across multiple query categories and baselines/versions, with validation of any LLM-judged metric against human labels \cite{liang2023helm}.

\subsubsection{Query Bank}\label{sec:query_bank}

We construct a query bank of $115$ science team questions, with the following categories:
\begin{itemize} 
\item \textbf{Compare potential new plan}: Add or replace points of interest or drill sites, force a visit order, shift the science station start time, restrict planning to an arbitrary region, run multi-target science requests, and inspect individual segments,
\item \textbf{Feasibility}: Evaluate confidence/risk of goal completion under probabilistic speed,
\item \textbf{Mission-document and source code lookup}: Retrieval-only questions answered from the mission documents, the planner code and outputs, or descriptions of the assistant's capabilities,
\item \textbf{Ambiguous/Edge Cases}: Prompts where the LLM should request clarification (e.g.,``is it safe?",``make it faster") or for impossible to answer queries (e.g.,``place a drill site outside of the bounds").
\end{itemize}

For each query, we record the LLM's expected behavior (answer, clarify, or refuse), the expected tools the human expert would use, any data files the LLM should reference, and the expected human response in operator vocabulary. 

\subsubsection{Testing Versions}\label{sec:variants}

We compare four LLM assistant versions to be tested on the same query bank, sharing the same underlying generator LLM, decoding parameters, and vocabulary guardrails. The four versions highlight the contributions of the selected scaffolding, tools, and retrieval in the complete version. The versions are: 
\begin{itemize} 
\item \textbf{Zero-shot LLM}: The generator receives only the user's query, with no system guardrails prompt, no tool catalog, and no retrieval layer,
\item \textbf{Closed-book LLM}: The generator receives the user query and the guardrails prompt, but no tools and no retrieval, so it only relies on prior knowledge,
\item \textbf{No-retrieval assistant}: ReAct with the action tools but without the retrieval layer,
\item \textbf{Full assistant}: All retrieval and action tools with ReAct.
\end{itemize}
We do not have a retrieval-only version with action tools removed because that chatbot could only answer lookup-related questions and not any plan comparison queries. 

\subsubsection{Success Metrics}\label{sec:metrics}
We organize our metrics into three categories. Trajectory metrics test task success on whether the LLM called the right tools with the right arguments. Extra tool calls and extra optional arguments are allowed, but every oracle tool and every oracle argument value must be present. We also report pass$^n$~\cite{yao2024tau}, the fraction of queries on which the assistant succeeds for all tested seeds. Task success is defined only for variants with tool access.
  
Semantic metrics test whether the natural-language response is accurate to the tool outputs and consistent with the human reference answer. For semantic accuracy, we report the fraction of factual claims in the response that are supported by the simulation data and tool outputs, ensuring the LLM is not favoring responding based on its prior knowledge. Because hand-labeling every response across all four versions and seeds is intractable at this scale, we evaluate responses on their semantic content with an LLM judge. We use a $1$--$5$ G-Eval \cite{liu2023geval} score that considers consistency, relevance, and vocabulary against the human reference and pairwise win rate \cite{zheng2023judge}, measuring which version's response the judge prefers when shown side by side. To validate the LLM judge, we hand-labeled a $50$-query subset on semantic accuracy (yes/no/uncertain) and behavior, and we report Cohen's $\kappa$ \cite{cohen1960kappa} between judge and human on each axis. Cohen's $\kappa$ is a standard agreement metric that measures how much a judge and a human agree beyond what would happen by chance, penalizing a judge that defaults to the dominant label in the dataset. Following common practice in LLM-judge validation \cite{tan2025judgebench}, we treat $\kappa < 0.6$ as a threshold below which we reject the judge's categorizations. The judge is selected from a different model family than the generator to reduce self-preference bias~\cite{panickssery2024self}.


Consistency metrics test whether semantically equivalent queries produce equivalent responses in the tools called and the semantic meaning. For queries whose expected behavior is to refuse or request clarification, we report whether the LLM executed the correct behavior. Each query is run independently through each variant for four seeds. We also note the monetary and latency costs of running each model. Reported numbers are means with $95\%$ bootstrap confidence intervals over the query bank, computed per query category. 

\section{Results}\label{sec:results}

%
%

\subsection{Planner Behavior}\label{sec:sim_validation}
\begin{figure*}[!t]
      \centering
      \captionsetup[subfigure]{skip=2pt}
      \newcommand{\panelw}{0.245\textwidth}
      \noindent\makebox[\textwidth]{%
        \hfill
        \begin{subfigure}[t]{\panelw}
        \vspace*{5pt}
            \includegraphics[width=\linewidth]{figures/example_trajectory.pdf}
            \caption{Original trajectory}\label{fig:sim-a}
        \end{subfigure}\hfill
        \begin{subfigure}[t]{\panelw}
        \vspace*{5pt}
            \includegraphics[width=\linewidth]{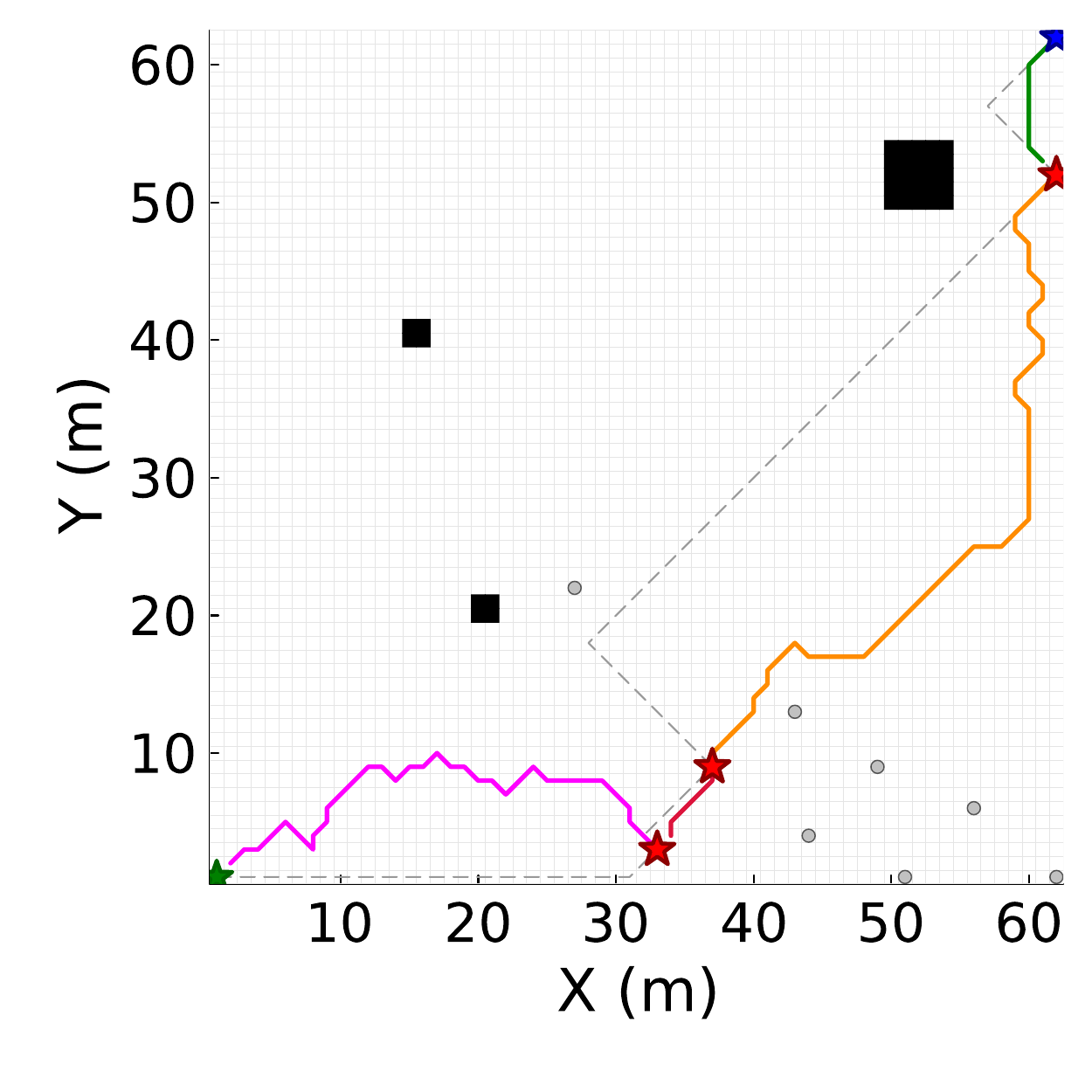}
            \caption{Start time shifted by $+12$\,h}\label{fig:sim-b}
        \end{subfigure}\hfill
        \begin{subfigure}[t]{\panelw}
        \vspace*{5pt}
            \includegraphics[width=\linewidth]{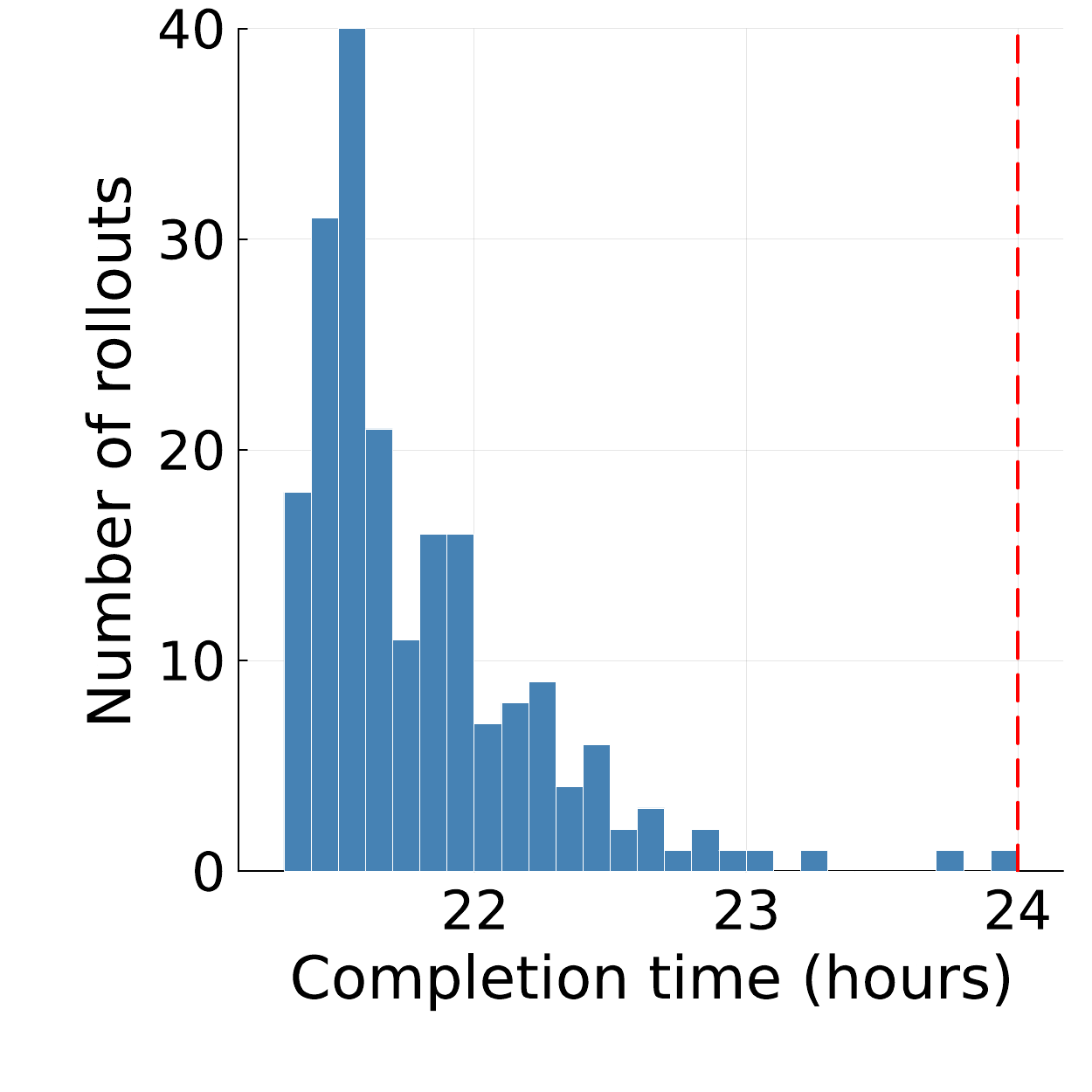}
            \caption{Completion-time risk}\label{fig:sim-d}
        \end{subfigure}\hfill
      }\par\vspace{6pt}
      \noindent\makebox[\textwidth]{%
        \null\hfill
        \begin{subfigure}[t]{\panelw}

            \includegraphics[width=\linewidth]{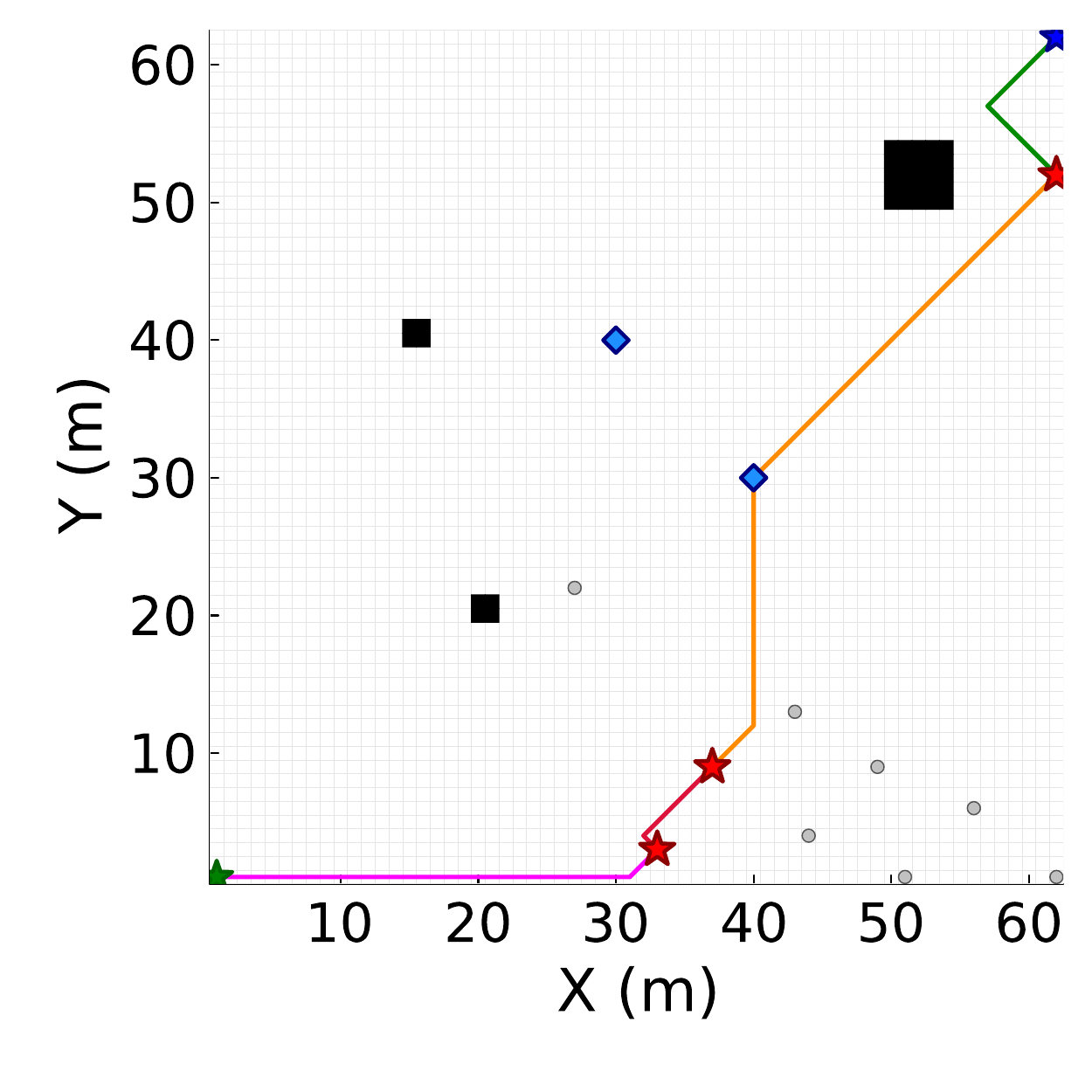}
            \vspace{-0.5cm}
            \caption{Two POIs added}\label{fig:sim-c}
        \end{subfigure}\hfill
        \begin{subfigure}[t]{\panelw}

            \includegraphics[width=\linewidth]{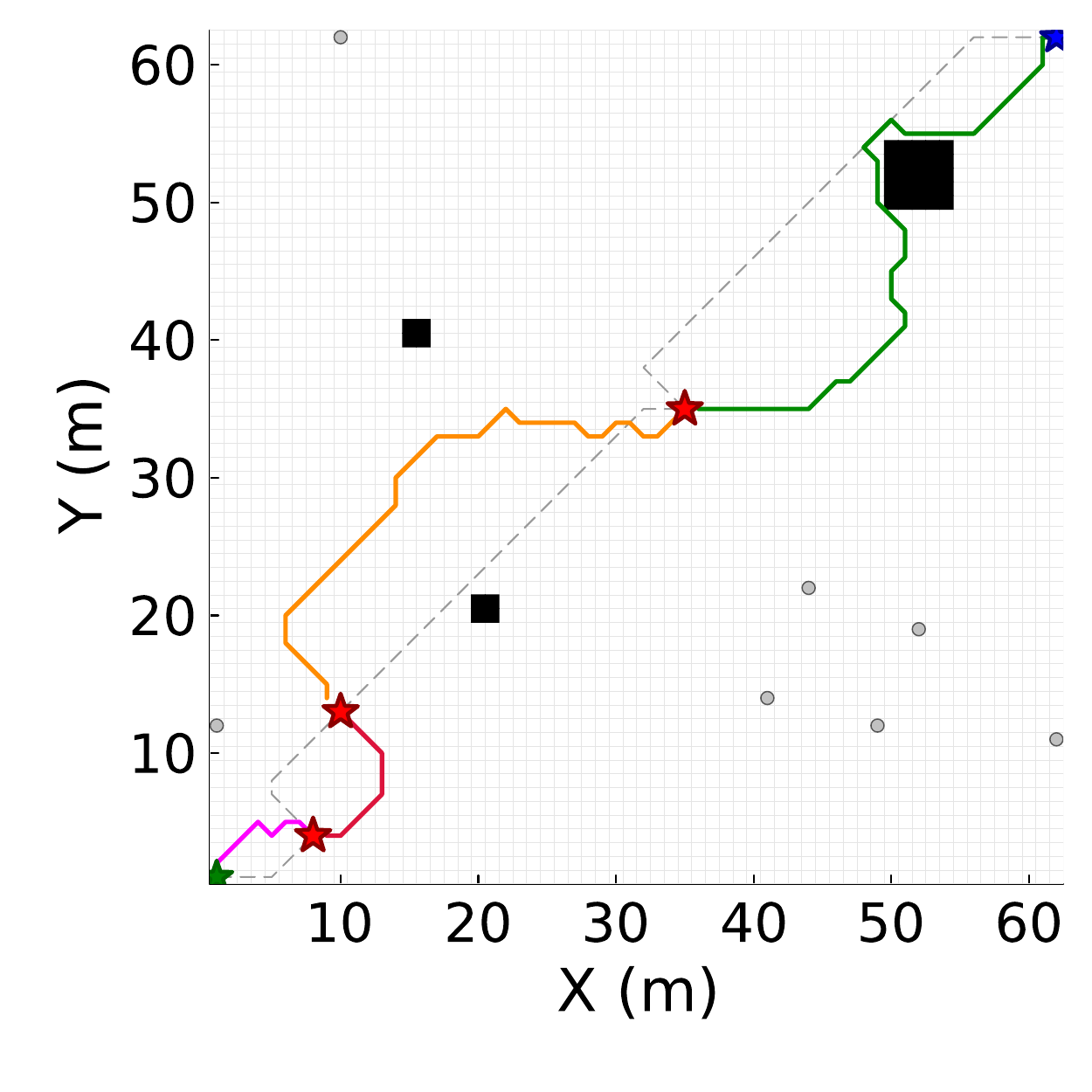}
            \vspace{-0.5cm}
            \caption{Alternate ice map}\label{fig:sim-e}
        \end{subfigure}\hfill
        \begin{subfigure}[t]{\panelw}

            \includegraphics[width=\linewidth]{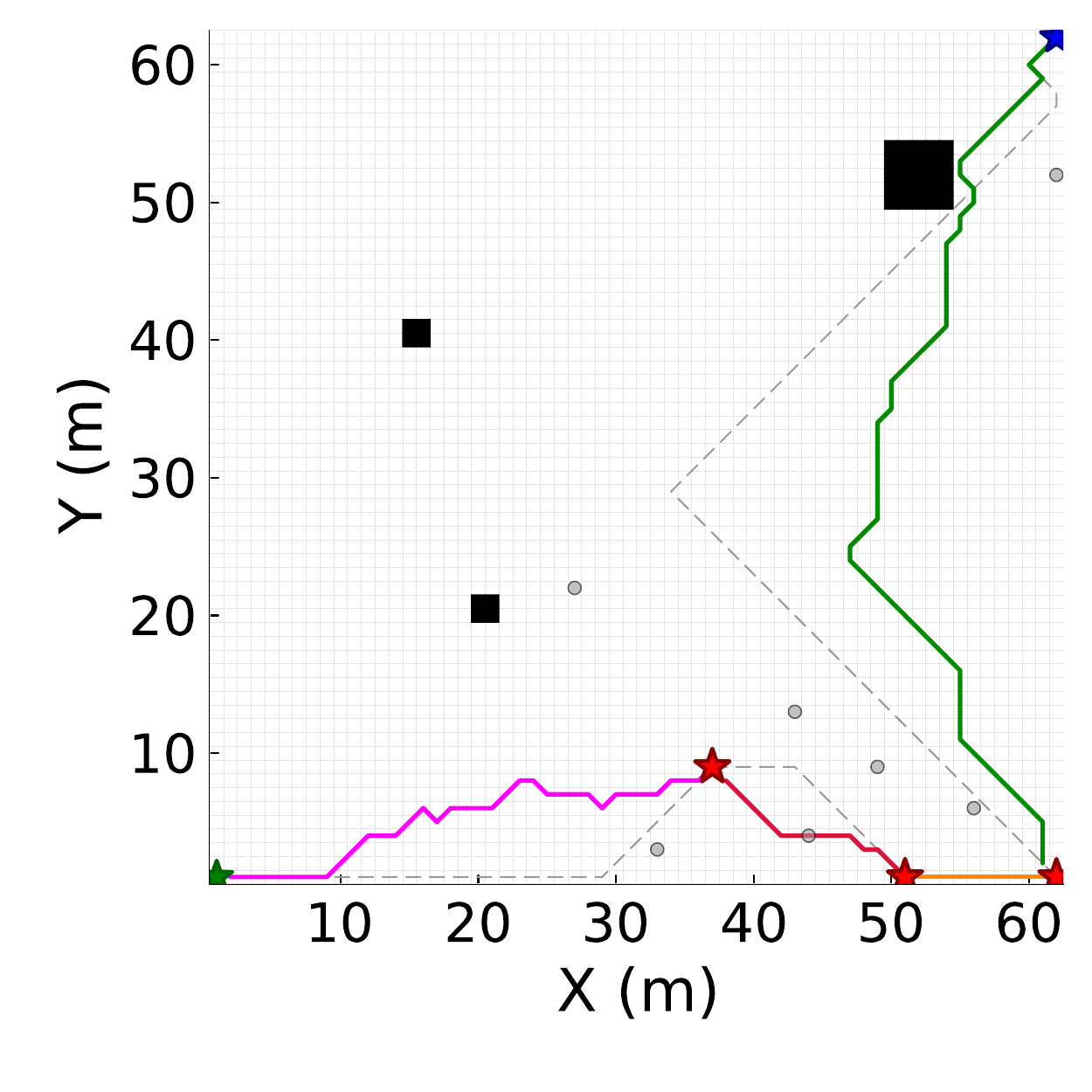}
            \vspace{-0.5cm}
            \caption{Drill order $3\rightarrow1\rightarrow2$}\label{fig:sim-f}
        \end{subfigure}\hfill
        \begin{subfigure}[t]{\panelw}
    
            \includegraphics[width=\linewidth]{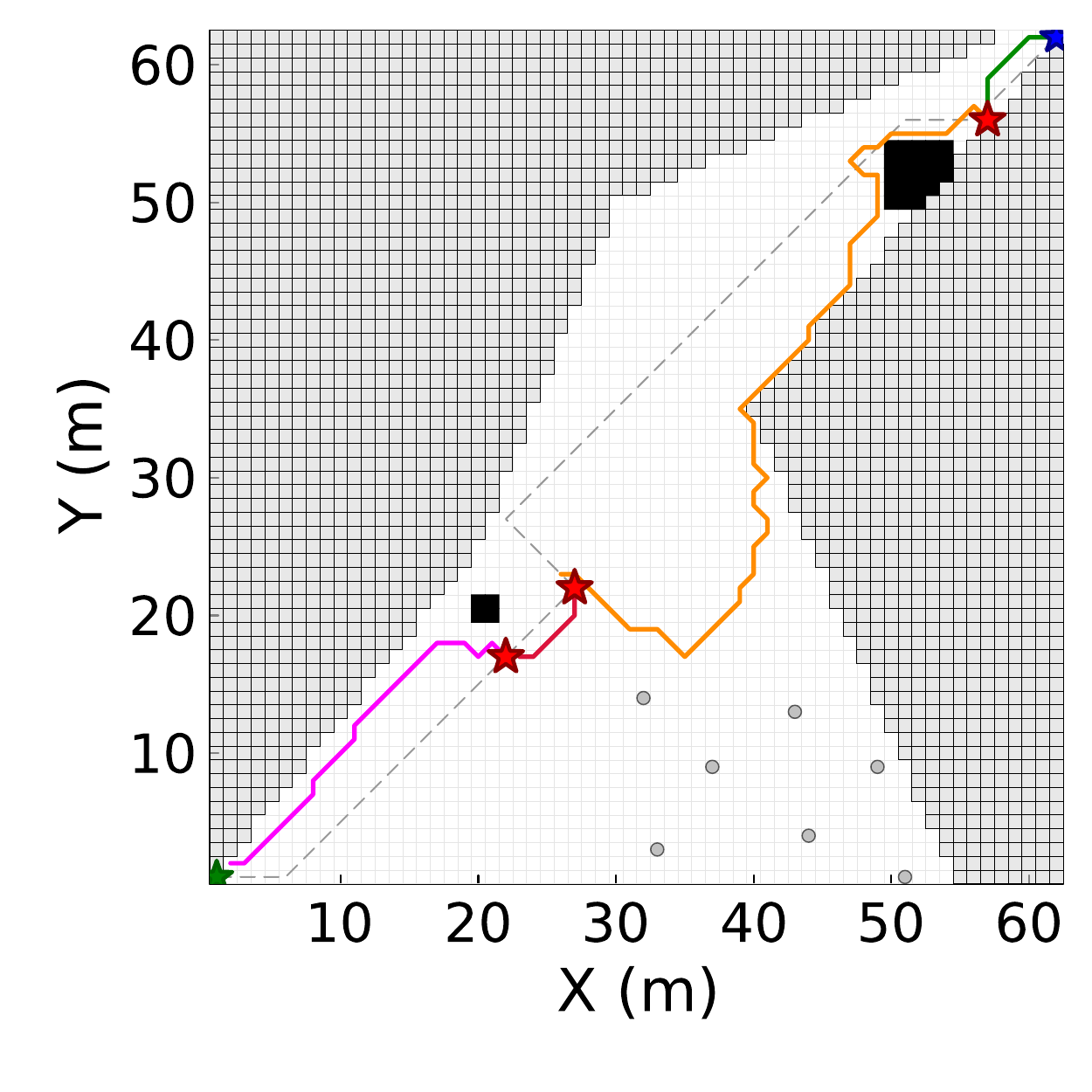}
            \vspace{-0.5cm}
            \caption{New science station area}\label{fig:sim-g}
        \end{subfigure} }
      \caption{Planner behavior demonstrated on different a. All panels share the same MDP formulation, number of drill sites selected from 10 candidates, and MCTS hyperparameters. They use the same legend from \Cref{fig:methodology-traj}, with blue diamonds as POIs. (a) The original plan takes 20.15 hours without prospecting. (b)~A $+12$\,h shift in start time results in a new plan due to different SMGs. (c)~200 Monte Carlo rollouts of (a) under SMG uncertainty, with 99.5\% finishing in 24 hours. (d)We added 2 POIs, finishing in 21.7\,h. (e)~We reran the planner with a new ice map. (f)~Enforcing a non-optimal site ordering increased the science station time to 24.45 hours, exceeding the limit. (g)~We replanned under new science station bounds.}
      \vspace{-0.3cm}
      \label{fig:sim-behavior}
  \end{figure*}

Before evaluating the framework, we verify that the underlying planner responds to the inputs that the science team may want to vary: start time, selected POIs, the underlying ice map, the site visit order, the boundary area, and the probabilistic SMG. \Cref{fig:sim-behavior} shows that each of these inputs changes the resulting plan in ways that match physical intuition. Delaying the start time by 12 hours changes both the route and the chosen drill set, since the planner might need to choose a different route to compensate for different SMG values. With our 200-rollout risk assessment, we observe that the science station durations cluster slightly above the deterministic estimate of 20.15 hours, with a small tail extending toward the 24-hour limit, a behavior consistent with the clipped SMG distribution. Across the rollouts, 99.5\% finished within the 24-hour limit. Inserting 2 POIs correctly changes the trajectory to route through. Replanning with a different ice map changes the drill candidates and chosen drill sites, therefore changing the trajectory. Enforcing a non-optimal drill order produced a 24.45-hour trajectory, exceeding the time limit, which was a reasonable result. Finally, we verified that the planner was able to plan within an arbitrarily defined science station area, given new start and end locations. Verifying that these results are reasonable for the planner,  we then evaluate the LLM assistant's ability to call the correct tools and information to explain the solutions.

\subsection{End-to-End Evaluation}\label{sec:results_main}

   \begin{table*}[!htbp]
      \renewcommand{\arraystretch}{1.15}
      \centering
      \caption{Results of chatbot/judge pairs for different framework versions across trajectory and semantic metrics on the query bank. Mean values are shown, and 95\% bootstrap confidence intervals are non-overlapping for every comparison reported. Trajectory metrics (task success, pass\textsuperscript{4}) are defined only for variants with a tool catalog.}
      \label{tab:variant-results}

      \scalebox{0.95}{
      \begin{tabular}{@{}l rrrr rrr@{}}
          \toprule
          & \multicolumn{4}{c}{\textbf{Chatbot: GPT-5.4 \quad Judge: Opus 4.8}}
          & \multicolumn{3}{c}{\textbf{Chatbot: Opus 4.8 \quad Judge: GPT-5.4}} \\
          \cmidrule(lr){2-5} \cmidrule(lr){6-8}
          \textbf{Metric}
          & \emph{zero shot} & \emph{closed book} & \emph{no retrieval} & \emph{full}
          & \emph{zero shot} & \emph{closed book} & \emph{full} \\
          \midrule
          \multicolumn{8}{@{}l}{\textit{Trajectory}} \\
          \quad task success            & ---           & ---  & 20.9 & \textbf{46.8} & ---  & ---  &
  \textbf{52.6} \\
          \quad pass\textsuperscript{n} & ---           & ---  & 19.8 & \textbf{41.9} & ---  & ---  &
  \textbf{48.8} \\
          \quad behavior acc.           & 78.0 & 33.7 & 50.0 & \textbf{79.1}          & 80.4 & 26.5 &
  \textbf{84.8} \\
          \midrule
          \multicolumn{8}{@{}l}{\textit{Semantic}} \\
          \quad faithfulness            & 26.7          & 56.1 & 72.6 & \textbf{90.9} & 23.4 & 10.1 &
  \textbf{65.3} \\
          \quad G-Eval (consistency)    & 48.2          & 47.8 & 51.8 & \textbf{59.8} & 47.3 & 45.6 &
  \textbf{60.4} \\
          \quad G-Eval (relevance)      & 75.2          & 65.1 & 71.6 & \textbf{85.1} & 80.9 & 59.6 &
  \textbf{88.2} \\
          \quad G-Eval (vocabulary)     & \textbf{79.2} & 64.3 & 68.5 & 75.8          & \textbf{79.6} & 59.5 & 70.5 
  \\
          \bottomrule
      \end{tabular}
      }
      \vspace{-0.5cm}
  \end{table*}

\Cref{tab:variant-results} reports the four-framework versions on the metrics defined in \Cref{sec:metrics}. Comparing against an expert-labeled dataset, trajectory metrics score whether the assistant selects the correct tools, documents, and behavior (answer, clarify, refuse), and semantic metrics score whether the response is accurate to the output trajectories and external documents provided. We use GPT-5.4 and Claude Opus 4.8 as they represent current frontier models from two distinct families. Across both chatbot/judge combinations of GPT and Opus, the full assistant (with the guardrail prompt, retrieval layer, and action tools) achieved the highest score on every metric except G-Eval (vocabulary), or how well the response reuses the science team's mission terminology. The zero-shot baseline, without tools or retrieval, can only ground its responses in the terminology from the query, whereas the other versions also bring in additional information surfaced by tool calls and retrieval. Vocabulary alone is not a comprehensive metric. The full framework outperforms on G-Eval (consistency) and (relevance), which measure whether the response is internally coherent and addresses the question asked, respectively. It also calls the correct tools with the correct arguments more often than the no-retrieval version, as shown with task success and pass\textsuperscript{n}, and had a higher behavior accuracy, though just barely above zero-shot, since most of the queries called for an answer, and zero-shot rarely refuses to answer or asks for clarification. The full version's responses are also preferred more often than the other versions' responses. For pairwise win rate, the Opus judge prefers the full assistant to closed-book 70.0\% of the time and to no-retrieval 65.4\%. The GPT judge preferred full to closed-book 85.2\% of the time. Moreover, as expected, the full version's responses were more correct than the other versions, as it had access to all tools and documents. 

We tested both combinations of chatbot/judge, with GPT-5.4 as chatbot and Opus 4.8 as judge, and vice versa. The full assistant outperforms on all metrics except G-Eval (vocabulary) for both combinations. Opus outperforms GPT on all metrics except for faithfulness. From a qualitative examination of the prompts, Opus's responses were more verbose, so even though there were correct facts, it often added fabricated details as well. For instance, for the query ``How much margin do we have under the 24-hour budget?" GPT correctly answered, "The current plan finishes in 20.15 hours, leaving 3.85 hours of margin under the 24-hour budget for prospecting," whereas Opus added a per-segment breakdown of slack and claimed that "the drill-3 segment contributed the largest share of the buffer at 1.4 hours," a number not given by any tool. Though Opus outperforms in other metrics, faithfulness is likely the most important metric. For the trajectory metrics, the assistant may use different tools than expected, but still get the correct response, and the G-Eval metrics are more valuable if accompanied by accurate responses. Moreover, Opus rarely refuses to answer or asks for clarification. The no-retrieval Opus version repeatedly tried to call inaccessible retrieval tools, exhausting tokens and time, and errored before producing a hallucinated response. 

  \begin{table}[H]
      \centering
      \vspace{-0.2cm}
      \caption{Cohen's $\kappa$ between LLM judge and author labels.}
      \footnotesize
      \begin{tabular}{@{}lcc@{}} 
      \toprule
      judge & $\kappa$ & agreement (\%) \\
      \midrule
      Opus 4.8        & $0.621$  & 87.8  \\
      GPT-5.4    & $-0.095$ & 50.0  \\
      \bottomrule
      \end{tabular}  
      \label{tab:kappa} 
      \vspace{-0.2cm}
  \end{table}
  
The judge's assessments are only trustworthy if they are aligned with a human expert's judgments. To validate the LLM judges, we hand-labeled a subset of 50 responses for faithfulness and behavior (answer/clarify/refuse). Calculating Cohen's $\kappa$ \cite{cohen1960kappa} as seen in \Cref{tab:kappa}, we see that Opus 4.8 is the better judge ($\kappa = 0.621$, with $87.8\%$ agreement between LLM judge and human preference) compared to GPT-5.4 ($\kappa = -0.095$ with $50.0\%$ agreement). Since  $\kappa_\text{opus} \geq 0.6$ \cite{tan2025judgebench}, we can rely on the judge's scores as a proxy for human-validated checks on the responses. Therefore, with this validation and high faithfulness score with GPT chatbot, we consider the better pairing to be GPT as the chatbot and Opus as the judge. This is further supported by the comparison of time and token monetary costs incurred by each model as chatbots. As seen in \Cref{tab:cost_latency}, we see that Opus requires much more time than GPT, almost three times as long. Future work will compare these expenditures against having the science team perform their validation. 

\begin{table}[ht]
\vspace{0.2cm}
    \centering
    \caption{Per-query chatbot latency in seconds}
    \vspace{-0.15cm}
    \footnotesize
    \begin{tabular}{@{}lrr@{}}
        \toprule
        Version & GPT-5.4 & Opus 4.8 \\
        \midrule
        zero\_shot    & 5.27  & 14.34  \\
        closed\_book  & 1.32  & 5.77   \\
        no\_retrieval & 9.02  & 170.46 \\
        full          & 10.83 & 29.77  \\
        \bottomrule
    \end{tabular}
    \vspace{-0.5cm}
    \label{tab:cost_latency}
\end{table}

Finally, to evaluate robustness to phrasing, we ran semantically equivalent queries on the full assistant. There was consistent accuracy between single-fact lookups (e.g., ``How long was the science station duration?"), specified a run with new conditions (e.g., ``Insert a POI at $(62,1)$"), or risk assessments (e.g., ``How confident are we that we'll make it to the goal on time?"). There were inconsistencies when the answer asked for just an explanation, not retrieving or running a tool. Near-synonyms such as asking for how much buffer, slack, or remaining margin returned the exact, correct value, but the assistant provided supporting math only for slack and margin. Queries like ``how does $X$ work?" or ``why does the path do $Y$?" produced inconsistent explanations. For example, asking variations of ``how does the planner weigh ice concentration vs. uncertainty?" incorrectly claimed that the planner prioritizes value and uses uncertainty as a tiebreaker or the correct formula. To address semantic inconsistencies, we can steer the LLM with DSPy prompt optimization \cite{khattab2024dspy}. The queries that failed faithfulness were on topics implemented in the planner, but had no grounding in retrieved documents. The full assistant's 90.9\% faithfulness is a nearly 65\% improvement over the 26.7\% zero-shot baseline. While hallucinations are a risk of any LLM system, we can improve through interactive sessions with the science team, addressing topics like tacking that are implemented but not discussed in the retrieved documentation.

\section{Conclusion}
We presented an LLM-based framework for explaining lunar rover trajectories planned under uncertainty, with an MCTS planner accompanied by a VIPER-inspired simulator that includes probabilistic SMGs, a 24-hour budget, and exploration vs. exploitation tradeoffs. The full framework with action tools and retrieval outperforms the zero-shot baseline, with a 90.9\% accuracy rate of responses. Our work has several opportunities for future improvement. We can extend the MDP to handle partial observability and incorporate active uncertainty reduction during trajectory planning. Additionally, we would like to test other solvers and use cases beyond MCTS and VIPER SST. The code repository for this work is
available at \url{https://michelleho-24.github.io/rover-explainablity/}.

\section*{Acknowledgments}
This work was supported by the NASA VIPER mission (Science Mission Directorate) and Amazon AI Fellowship. 

\vspace{-0.1cm}
\bibliographystyle{IEEEtran}
\bibliography{references}

\end{document}